\documentclass[sigconf,natbib=true]{acmart}
\usepackage{amsmath,amsfonts}
\usepackage{adjustbox}

\usepackage{algorithm}
\usepackage[noend]{algpseudocode}
\usepackage{graphicx}
\usepackage{textcomp}
\usepackage{xcolor}
\usepackage{caption}
\usepackage{subcaption}
\usepackage{tabularx}
\usepackage{multirow}
\usepackage{makecell}
\usepackage{diagbox}
\algrenewcommand\algorithmicrequire{\textbf{Input:}}
\algrenewcommand\algorithmicensure{\textbf{Output:}}

\AtBeginDocument{
  \providecommand\BibTeX{{
    \normalfont B\kern-0.5em{\scshape i\kern-0.25em b}\kern-0.8em\TeX}}}

\setcopyright{acmcopyright}
\copyrightyear{2022}
\acmYear{2022}

\author{Yoonhyuk Choi}
\orcid{0000-0003-4359-5596}
\affiliation{
  \institution{
  Seoul National University}
  \city{Seoul}
  \state{Republic of Korea}
}
\email{younhyuk95 @ snu.ac.kr}

\author{Jiho Choi}
\orcid{0000-0002-7140-7962}
\affiliation{
  \institution{
  Seoul National University}
  \city{Seoul}
  \state{Republic of Korea}
}
\email{jihochoi @ snu.ac.kr}

\author{Taewook Ko}
\orcid{0000-0001-7248-4751}
\affiliation{
  \institution{
  Seoul National University}
  \city{Seoul}
  \state{Republic of Korea}
}
\email{taewook.ko @ snu.ac.kr}

\author{Hyungho Byun}
\orcid{0000-0003-1908-637X}
\affiliation{
  \institution{
  Seoul National University}
  \city{Seoul}
  \state{Republic of Korea}
}
\email{notorioush2 @ snu.ac.kr}

\author{Chong-Kwon Kim}
\orcid{0000-0002-9492-6546}
\affiliation{
  \institution{Korea Institute of Energy Technology}
  \city{Naju}
  \state{Republic of Korea}
}
\email{ckim @ kentech.ac.kr}

\renewcommand{\shortauthors}{Yoonhyuk Choi et al.}

\acmConference[CIKM '22] {Proceedings of the 31st ACM International Conference on Information and Knowledge Management}{October 17--21, 2022}{Atlanta, GA, USA.}
\acmBooktitle{Proceedings of the 31st ACM International Conference on Information and Knowledge Management (CIKM '22), October 17--21, 2022, Atlanta, GA, USA}
\acmPrice{15.00}
\acmISBN{978-1-4503-9236-5/22/10}
\acmDOI{10.1145/3511808.3557324}

\begin{document}

\title{Finding Heterophilic Neighbors via Confidence-based Subgraph Matching for Semi-supervised Node Classification}

\begin{abstract} 
Graph Neural Networks (GNNs) have proven to be powerful in many graph-based applications. However, they fail to generalize well under heterophilic setups, where neighbor nodes have different labels. To address this challenge, we employ a confidence ratio as a hyper-parameter, assuming that some of the edges are disassortative (heterophilic). Here, we propose a two-phased algorithm. Firstly, we determine edge coefficients through subgraph matching using a supplementary module. Then, we apply GNNs with a modified label propagation mechanism to utilize the edge coefficients effectively. Specifically, our supplementary module identifies a certain proportion of task-irrelevant edges based on a given confidence ratio. Using the remaining edges, we employ the widely used optimal transport to measure the similarity between two nodes with their subgraphs. Finally, using the coefficients as supplementary information on GNNs, we improve the label propagation mechanism which can prevent two nodes with smaller weights from being closer. The experiments on benchmark datasets show that our model alleviates over-smoothing and improves performance.
\end{abstract}

\ccsdesc[500]{Computing methodologies~Machine learning}

\keywords{Graph Neural Networks, Graph Representation Learning, Graph Similarity Computation, Label Propagation}

\maketitle

\section{Introduction}
The investigation of graph-structured data has gained significant attention in various fields; physics \cite{gilmer2017neural}, protein-protein interactions \cite{fout2017protein}, and social networks \cite{fan2019graph}. Integrated with deep neural networks (DNNs) \cite{lecun2015deep}, graph neural networks (GNNs) have achieved state-of-the-performance by concurrently modeling node features and network structures \cite{scarselli2008graph,defferrard2016convolutional,kipf2016semi,hamilton2017inductive,velickovic2017graph}. Specifically, the message passing plays an important role by aggregating features from neighboring nodes \cite{gilmer2017neural}. Consequently, GNNs often have shown the best performance in various tasks including semi-supervised node classification and link prediction.

However, recent studies reveal that GNNs gain advantages of message passing under limited conditions, e.g., high assortativity of subject networks \cite{mcpherson2001birds}. In this paper, we assume two types of networks; homophilic (assortative) ones where most edges connect two nodes with the same label, and heterophilic graphs where the most connections are disassortative. Most prior work on GNNs assumes that connected nodes are likely to possess the same label, and thus, they fail to attain sufficient performance for many real-world heterophilic datasets \cite{pandit2007netprobe}. Many clever schemes have been introduced to solve the problem. Some of them specify different weights for each connection \cite{velickovic2017graph,yang2019masked,bo2021beyond,kim2022find}, or remove disassortative edges \cite{ying2019gnnexplainer,entezari2020all,luo2021learning}. Others employ distant nodes with similar features \cite{pei2020geom,yang2021graph,jin2021node} or apply different aggregation boundary based on the central nodes \cite{xiao2021learning}. Nonetheless, there is a question to be addressed: \textit{is it necessary to specify different weights for GNNs?}

\begin{figure*}
     \centering
     \begin{subfigure}[b]{0.32\textwidth}
         \centering
         \includegraphics[width=1.1\textwidth]{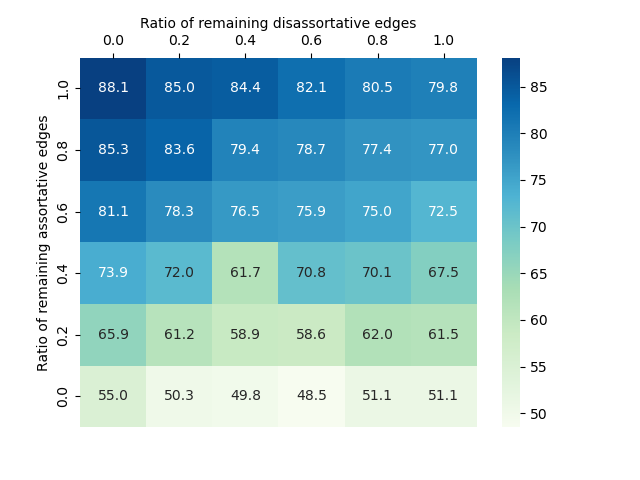}
         \caption{}
         \label{example_2_a}
     \end{subfigure}
     \hfill
     \begin{subfigure}[b]{0.32\textwidth}
         \centering
         \includegraphics[width=1.1\textwidth]{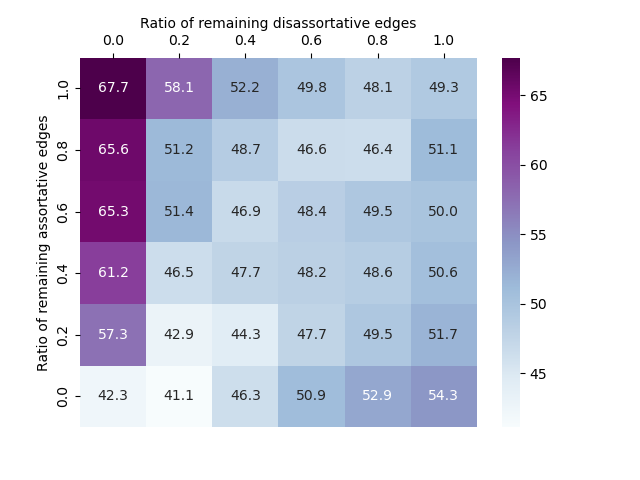}
         \caption{}
         \label{example_2_b}
     \end{subfigure}
     \hfill
     \begin{subfigure}[b]{0.32\textwidth}
         \centering
         \includegraphics[width=.8\textwidth]{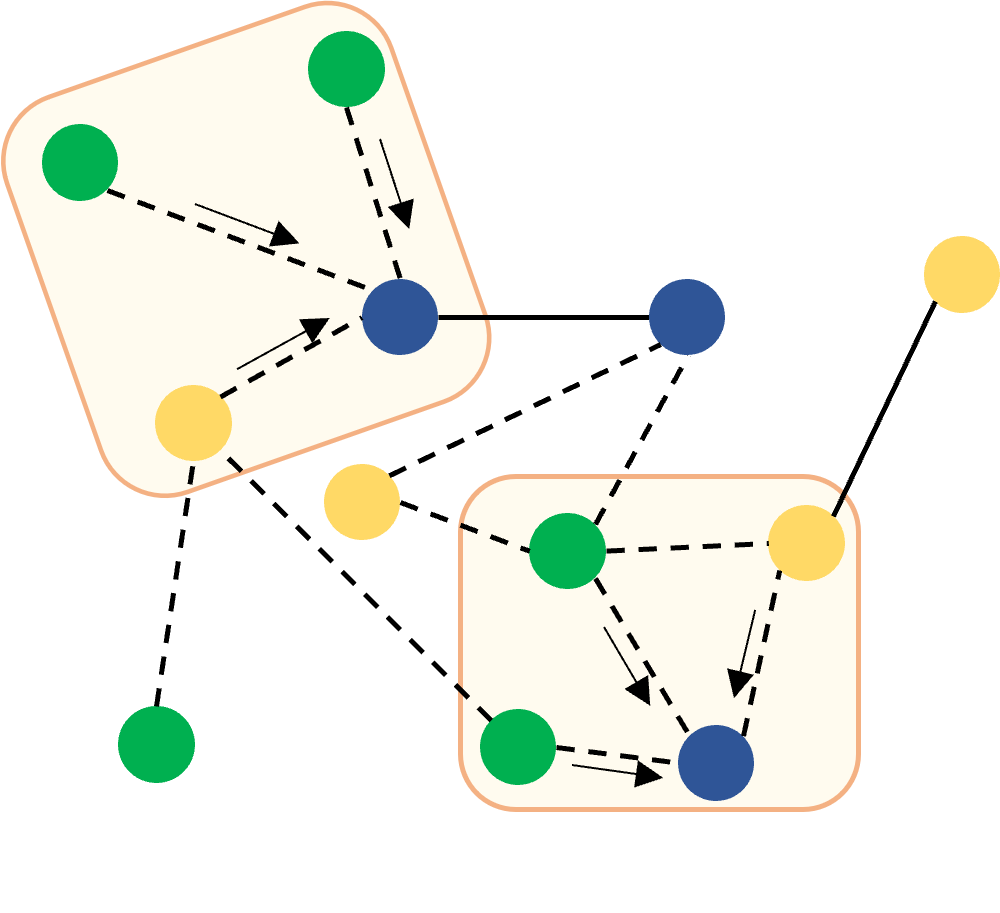}
         \caption{}
         \label{example_2_c}
     \end{subfigure}
        \caption{Node classification accuracy (\%) of GCN on different datasets; (a) Cora, and (b) Chameleon. For each graph, we randomly prune a certain proportion of assortative / disassortative edges and plot their performance. We also describe a special case of (c) helpful aggregation scenario under disassortative graphs}
        \label{example_2}
\end{figure*}

To answer the above question, we conduct an investigation using two representative datasets; one is an assortative citation network called Cora \cite{mccallum2000automating}, and the other one is Chameleon \cite{rozemberczki2019gemsec} which contains many disassortative links between Wikipedia web pages. In Figure \ref{example_2}, we randomly prune a certain ratio of assortative / disassortative edges and describe the node classification accuracy of GCN \cite{kipf2016semi}. Through this study, we observe two characteristics; (1) for Cora, the performance increases as the assortative edges are maintained, while disassortative edges are removed. On the contrary, Chameleon data is rather heterophilic and thus, the disassortative links play an important role as the number of remaining assortative edges becomes smaller. To analyze this result, we take Figure \ref{example_2}-c as an example. Though the graph is heterophilic, two central nodes share the same types of neighborhoods (2 green, 1 yellow) that can contribute to distinguishing them from others \cite{luan2021heterophily,ma2021homophily}. (2) the removal of assortative edges has a greater impact on the overall performance than disassortative ones. For example, the performance in Cora using the original graph is 79.8 \% (top right). If we remove all disassortative edges, it attains 88.1 \% (top left), whereas eliminating assortative links becomes 51.1 \% (bottom right). 
To summarize, we conjecture
that removing a small proportion of assortative edges can be harmful, and thus, assigning accurate weights are fundamental for GNNs. Now, the problem is; \textit{how can we figure out these coefficients correctly and utilize them?}

To achieve this, we focus on the GAM \cite{stretcu2019graph} that suggests a supplementary module with label propagation. Specifically, the supplementary module of GAM only utilizes a central node to debilitate noises. However, referring to Figure \ref{example_2}, excluding all links of assortative and disassortative shows the lowest performance, which is the same as GAM's method. 
To solve this limitation, our supplementary module focuses on the widely used optimal transport \cite{peyre2019computational,xu2019gromov,mialon2020trainable,kolouri2020wasserstein} to
measure similarity between two subgraphs. In addition, we further apply a confidence ratio to deal with multiple disassortative links. 
Then, considering these predictions as supplementary edge coefficients, we apply label propagation \cite{bui2018neural} between a certain proportion of high confident edges, while the others are considered disassortative and the connected nodes are prevented from being similar.
Our contributions can be summarized as follows:

\begin{itemize}
\item We introduce a confidence-based subgraph matching to retrieve edge coefficients accurately. Our model is scalable and generalizes well for both homophilic / heterophilic graphs, which can be achieved by varying the values of the confidence ratio.
\item Assuming that a certain proportion of entire edges are disassortative, we improve the label propagation to keep two nodes with a lower similarity score from being closer. Specifically, we divide the edge coefficients into two parts, which can guide the positive pairs to be similar and vice versa. 
\item We conduct extensive experiments on publicly available datasets to validate the above suggestions. The ablation studies indicate the superiority of subgraph matching techniques for retrieving class sharing probability. 
\end{itemize}

\section{Related Work}
Graph neural networks (GNNs) have shown substantial improvement for semi-supervised classification tasks. Most of them can be categorized into two types; spectral-based and spatial-based methods. The first one utilizes structural information of the entire graph through Laplacian decomposition \cite{hammond2011wavelets} that requires high computational costs $O(n^3)$. To reduce their complexity, GCN \cite{kipf2016semi} suggests a first-order approximation of Chebyshev polynomials \cite{defferrard2016convolutional} and utilizes features of neighboring nodes by simply stacking convolutional layers. Ada-GNN \cite{dong2021graph} further employs an adaptive frequency filter to capture different perspectives of nodes. 
However, these algorithms inevitably aggregate noisy adjacent nodes, where they assume two connected nodes are likely to share the same label. 

Recently, some algorithms focus on the retrieval of edge coefficients using the node features. For example, GAT \cite{velickovic2017graph} measures the relevance between two nodes by applying an attention layer to their features. Similarly, Masked-GCN \cite{yang2019masked} estimates attribute-wise similarity for precise propagation. GNNExplainer \cite{ying2019gnnexplainer} identifies the set of important edges and features that maximize the mutual information of the final prediction. Nonetheless, these methods may fail to generalize well under a heterophilic graph, where the message passing inevitably makes two connected nodes similar. 

To solve this problem, FAGCN \cite{bo2021beyond} selects whether to propagate low-frequency or high-frequency signals by enabling edges to have negative coefficients. L2Q \cite{xiao2021learning} parameterizes the aggregation boundary of each node to deal with heterophily.
SuperGAT \cite{kim2022find} differentiates between friendly and noisy neighbors based on their homophily and node degrees. However, these methods also implicate noisy information since they work as a downstream task of GNNs.
Some argue that graph sparsification \cite{zheng2020robust,desai2021graph} is considerable for graph denoising. For example, PTDNet \cite{luo2021learning} adopts nuclear norm to prune edges between communities. Yet, it also implicates risk for pruning positive edges and is not powerful enough for classification compared to classical GNNs.

As another branch, non-local neural networks \cite{wang2018non,liu2021non} have gained increasing attention for capturing long-range dependencies. Since previous GNNs only utilize local adjacent nodes, they fail to deal with heterophilic graphs.
Instead of directly specifying coefficients, finding distant but similar nodes has increased the representational power of GNNs. Specifically, Geom-GCN \cite{pei2020geom} further exploits distant nodes within a specific boundary, and executes grid-based aggregation. Simp-GCN \cite{jin2021node} mixes the original adjacency matrix with a feature-based similarity matrix through learnable parameters. Nonetheless, they implicate two limitations.
Firstly, operating as a downstream task of GNNs may inevitably contain noisy information after aggregation. Secondly, measuring relevance between two nodes can be biased (or risky) under a semi-supervised setting that has few labeled samples \cite{liu2022confidence}.

Apart from retrieving edge coefficients, a strategy for utilizing this information is also considerable. For example, P-reg \cite{yang2021rethinking} simply utilizes entire edges to provide additional information for GNNs.
NGM \cite{bui2018neural} integrates label propagation (LP) with GNNs, while GAM \cite{stretcu2019graph} further parameterize edge coefficients. 
However, these methods are highly localized and fail to discriminate less important edges under the global aspect. Further, they show limited performance for precise prediction under our experiments.   
Instead, we focus on pairwise matching between two subgraphs that are independent of GNN modules. Using the mechanism of optimal transport (OT) \cite{xu2019scalable,kolouri2020wasserstein,mialon2020trainable}, we integrate a confidence-based denoising network to secure robustness, followed by our label propagation.

\section{Notations}
We first set up the problem with the commonly used notations of graph-structured data. Let us define an undirected graph $G=(\mathcal{V},\mathcal{E})$, where $\mathcal{V}=\{v_1, v_2, ..., v_N\}$ represents set of $N$ nodes and $\mathcal{E}=\{e_1, e_2, ..., e_M\}$ denotes the set of $M$
edges. $A \in \mathbb{R}^{N \times N}$ stands for adjacency matrix of graph $G$. 
The properties of entire nodes $N$ are represented as feature matrix $X \in \mathbb{R}^{N \times F}$ with dimension $F$. We assume the label matrix $Y \in \mathbb{R}^{N \times C}$, where $C$ is the number of classes. Each row of $Y$ corresponds to a one-hot vector of node $v$'s label. Using partially labeled nodes $\mathcal{V}_L \subset \mathcal{V}$ as a training set, we aim to infer the labels of unlabeled nodes $\mathcal{V}_U=\mathcal{V}-\mathcal{V}_L$.

\section{Methodologies}
\begin{figure*}[ht]
    \includegraphics[width=\textwidth]{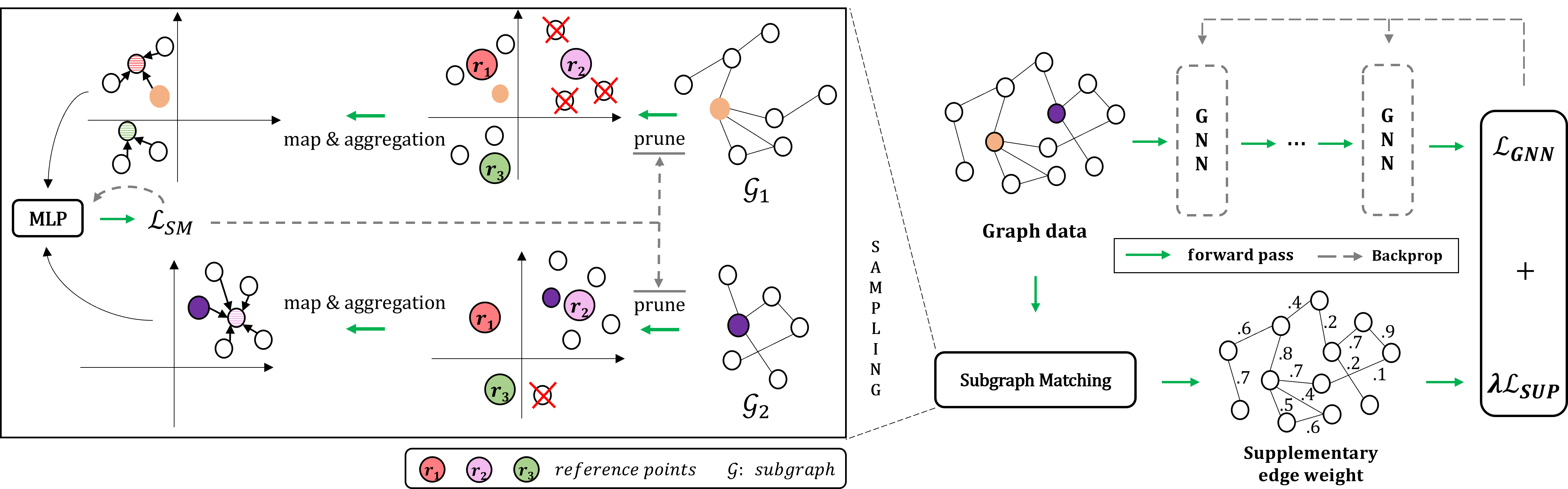}
    \caption{The overall framework of our model. It consists of two parts; one for the subgraph matching module which generates supplementary edge coefficients, and the other one is the GNN module that utilizes weights for label propagation}
  \label{model}
\end{figure*}

Figure \ref{model} illustrates the overall architecture of our model which consists of two parts. On the right side, we describe the GNN module with label propagation which takes the predicted edge weights for training. The left one stands for the subgraph matching that provides edge coefficients as supplementary information.

The two modules do not share loss or parameters and are updated independently. In Section \ref{edge_coef}, we first introduce methodologies for retrieving edge coefficients, followed by our subgraph matching module. In Section \ref{label_prop}, we suggest strategies to utilize these predictions effectively through label propagation.

\subsection{Retrieving Edge Coefficients} \label{edge_coef}
Recently, many efforts have been dedicated to specifying edge coefficients, and we categorize them into two types. Firstly, in Section \ref{puc}, we take previous methods that only utilize central nodes for classification. Secondly, in Section \ref{pus}, we describe previous algorithms that further utilize the adjacent nodes for prediction. Finally, we discuss the advantages and limitations of these methods and describe our subgraph matching module in Section \ref{suggest}.

\subsubsection{Retrieving edge coefficients using a central node} \label{puc}
These types of methods include message passing, but only a central node is used for similarity measure, not a subgraph. With the slight abuse of notation, let us assume the $h_i,h_j$ as hidden representations of two nodes $i,j$. 

\textbf{Graph agreement model (GAM)} \cite{stretcu2019graph} introduces an auxiliary model to predict a same class probability $w_{ij}$ between two nodes $i,j$ as below:
\begin{equation}
\label{gam}
w_{ij}=MLP ((h_i-h_j)^2).
\end{equation}
The $MLP$ is a fully-connected network with non-linear activation. GAM works well under the heterophilic graph since they do not utilize neighboring nodes. However, as the homophily ratio of the graph increases, we notice that they show significantly lower performance even compared to the plain GCN \cite{kipf2016semi}.

\textbf{Graph attention network (GAT)} \cite{velickovic2017graph} applies layer-wise attention as a downstream task of GNN as below:
\begin{equation}
w^l_{ij}={exp(\sigma({a^l}[h^l_i||h^l_j])) \over \sum_{k \in N_i}exp(\sigma({a^l}[h^l_i||h^l_k]))}.
\end{equation}
GAT specifies different weights for each layer, where $a^l$ is a learnable vector at the $l-th$ layer. Compared to GAM, a softmax function normalizes the weights that are highly dependent on the degree of each node, which makes it harder to determine their importance. Further, the message passing can degrade the performance since the edge coefficients $w_{ij}$ always maintain a positive value.

\textbf{FAGCN} \cite{bo2021beyond} improves GAT from two perspectives; replacing softmax with degree-based normalization, and adopting different activation function as below: 
\begin{equation}
\label{fagcn}
w^l_{ij}=tanh(a^l[h^l_i||h^l_j]).
\end{equation}
The main difference lies in $tanh$, where the negative value of coefficients can maintain high-frequency signals. However, we notice that their accuracy decreases as the homophily of networks increases (e.g., Cora), where all coefficients converged to a positive value and fail to figure out heterophilic edges. 

\textbf{PTDNet} \cite{luo2021learning} removes task-irrelevant edges by applying randomness $\epsilon$ and decaying factor $\gamma$. Here, the coefficients $w_{ij}$ can be derived as below:
\begin{equation}
w^l_{ij}=\sigma((\log\epsilon^l - \log(1-\epsilon^l)+MLP(h^l_i,h^l_j))/\gamma).
\end{equation}
The random value follows $\epsilon^l \sim Uniform(0, 1)$, and decaying factor $\gamma$ depends on the iteration number. They apply nuclear norm on the entire edges $w$ to remove connections between communities. However, we notice that randomness can impede precise prediction, and nuclear norm does not always lead to optimal results.

Summarizing the above methodologies, prediction based on the central node implicates two major problems. Firstly, excluding message passing (GAM) can lead to over-fitting, where it contains limited information. Though other methods incorporate neighboring nodes, the noisy neighbors also participate in the aggregation process, which can impede robustness and incur over-smoothing issues \cite{zhao2019pairnorm}.
Secondly, directly employing the coefficients as an adjacency matrix is highly risky, where the elimination of assortative edges hurts the overall performance of GNNs (please refer to Figure \ref{example_2}). To solve these limitations, we focus on subgraph matching algorithms which will be introduced in the upcoming section.

\subsubsection{Retrieving edge coefficients using subgraphs} \label{pus}
In this section, we describe some methods of measuring the similarity between two subgraphs. Recently, applying optimal transport (OT) on subgraphs \cite{peyre2019computational,xu2019gromov} has shown great improvement, which is a mathematical framework for measuring distances (similarity) between objects. For example, let us assume two subgraphs $\mathcal{G}_i,\mathcal{G}_j$ that contain $m,n$ nodes, respectively. Then, we can define transport (coupling) matrix $P \in \mathcal{R}^{m \times n}$ between two subgraphs that meets $P1_n={1 \over m}1_m$ and $P^T1_m={1 \over n}1_n$. The objective of OT is to find matrix $P$ that minimizes the function below:
\begin{equation}
\label{sinkhorn}
\min_P\sum_{ij}{S(\mathcal{G}_i,\mathcal{G}_j)P_{ij}} - \epsilon H(P).
\end{equation}
Here, $S$ is a cost function and $H(\cdot)$ is entropy regularized Kantorovich relaxation with regularizer $\epsilon$. However, finding $P_{ij}$ for all pairs of $(i, j)$ requires a high computational cost. 

\textbf{Linear optimal transport} \cite{mialon2020trainable} employ reference points $r$ to solve the above limitation, which can be retrieved through $k$-means clustering or calculating Wasserstein barycenter \cite{cuturi2014fast} based on each class of training nodes. Here, elements that are assigned to the same cluster (reference) are pooled together and thus, reducing the pair-wise calculation.
Specifically, matrix $P \in \mathcal{R}^{C \times N}$ splits or assigns the entire node $N$ to references $r$ ($C$ stands for the number of reference points). $P$ can be obtained through multiple ways (e.g., Sinkhorn's algorithm \cite{sinkhorn1967concerning}), which calculates a relevance between the inputs and reference points as below:
\begin{equation}
\label{optimal_transport}
p^*_i=\mathop{\arg\min}_{p \in P_i}\sum_{k=1}^{C}\sum_{l=1}^{N_i}p_{kl}\Vert r_k-h_l \Vert^2
\end{equation}
, where $N_i$ is the number of nodes in subgraph $i$. 
Let us assume the hidden representations of two subgraphs as $h_i \in \mathcal{R}^{N_i \times F},h_j \in \mathcal{R}^{N_j \times F}$ whose feature dimension is $F$. Using $p^*_i \in \mathcal{R}^{C \times N_i}$, $p^*_j \in \mathcal{R}^{C \times N_j}$ in Equation \ref{optimal_transport} that splits the mass of subgraph $h_i,h_j$ to multiple references $h'_i,h'_j \in \mathcal{R}^{C \times F}$, one can measure their similarity through matching function M (MLP) as below:
\begin{equation}
\label{linear_ot}
\begin{gathered}
h'_i= p^*_ih_i, \,\,h'_j= p^*_jh_j \\
w_{ij}=M(h'_i,h'_j).
\end{gathered}
\end{equation}

\textbf{Monge map} \cite{kolouri2020wasserstein} does not split mass, while an injective mapping is applied for each subgraph as follows:
\begin{equation}
\label{monge_map}
\begin{gathered}
h'_i = B(p^*_i, h_i), \,\,h'_j = B(p^*_j, h_j) \\
w_{ij}=M(h'_i,h'_j).
\end{gathered}
\end{equation}
Similar to linear optimal transport \cite{mialon2020trainable}, each subgraph $h_i,h_j$ can be mapped to new points $h'_i,h'_j \in \mathcal{R}^{C \times F}$ through optimal transport $p^*$ (please refer to Eq. \ref{optimal_transport}). 
The difference lies in a barycentric projection $B$ that ensures no mass splitting (please read this paper \cite{kolouri2020wasserstein} for more details). 

Using the insight of these methods \cite{mialon2020trainable,kolouri2020wasserstein}, the subgraph matching has the advantage of using adjacent nodes for predictions. However, they also implicate a limitation of handling noisy neighbors, since they utilize the entire nodes of the subgraph to measure their similarity. To deal with this, we now introduce our method that utilizes a confidence ratio as below.

\subsubsection{\textbf{Our subgraph matching using a confidence ratio.}} \label{suggest} 
In Figure \ref{model}, we describe the overall architecture of our \underline{Con}fidence-based \underline{S}ubgraph \underline{M}atching. ConSM calculates a similarity between two subgraphs (probability of sharing the same label) using optimal transport and confidence ratio as follows:
\begin{enumerate}
    \item Sampling: We randomly sample two labeled nodes, whose labels can be the same or different. 
    For sampling, the size of the positive and negative pairs should be the same to avoid a model being biased. We further utilize their 2-hop adjacent nodes as inputs.
    \item Prune: We measure a score of entire edges through reference points. Then, based on a confidence ratio, we maintain top-$k$ confident ones while removing others.
    \item Map and aggregation: Given two subgraphs $\mathcal{G}_i,\mathcal{G}_j$, we first map nodes to low-dimensional embedding $h_i,h_j$ and assign them to the nearest reference points $r$ through the Monge map. Then, we aggregate the nodes that belong to the same reference points by pooling operation (e.g., mean).
    \item Prediction: We measure the similarity of the two graphs and also retrieve the class probability of a central node. 
\end{enumerate}
Now, we describe the details of our method below.

\textbf{Sampling}\label{sampling} We adopt an auxiliary module for retrieving edge coefficients
that are independent of the GNN module. Here, two nodes are randomly sampled based on their class. If two nodes share the same class (positive pair), we assume the label of this pair as 1 and otherwise 0. To prevent a class imbalance problem, the same number of positive and negative pairs are sampled. Compared to GAM \cite{stretcu2019graph}, we utilize the subgraph of a central node (adjacent nodes within 2-hop) to improve prediction accuracy.

\textbf{Prune}
Unlike previous method \cite{kolouri2020wasserstein} that applied \textit{embedding $\rightarrow$ aggregation $\rightarrow$ mapping}, we suggest \textit{embedding $\rightarrow$ pruning $\rightarrow$ mapping $\rightarrow$ aggregation}' to handle noisy edges. 
Specifically, we only utilize a certain proportion of edges based on their scores and a confidence ratio ($\zeta$).
We first describe our scoring function. Using the initial node features $X$, we can retrieve their low-dimensional embedding $h \in \mathcal{R}^{N \times F}$ through an encoder (MLP) as:
\begin{equation}
\label{embedding}
h = Encoder(X).
\end{equation}
Similarly, the embedding of reference points is $r \in \mathcal{R}^{C \times F}$, which can be obtained through class-wise averaging of training nodes. Then, we can measure a score (e.g., cosine similarity) $S \in \mathcal{R}^{N \times C}$ between nodes $h$ and references $r$ as below:
\begin{equation}
\label{similarity}
S = h \cdot r^T
\end{equation}
, where the row of $S$ represents a score of each node with respect to the reference points. Given two nodes $i,j$, we can retrieve their similarity $w_{ij}=S_i \cdot S_j$. Consequently, the $w$ of the entire edges can be obtained, and thus, we manage to maintain top-$k$ edges $k=\lfloor \zeta \times |\mathcal{E}|\rfloor$ while removing others.

\textbf{Map and aggregation}
Using the remaining edges, the adjacency matrix can be reconstructed. With the slight abuse of notation, given two nodes $i,j$ and their subgraphs $\mathcal{G}_i,\mathcal{G}_j$, we assume that their subgraph embedding $h_i,h_j$ can be retrieved as below: 
\begin{equation}
\label{feat_trans}
h_i = Encoder(\mathcal{G}_i), \,\, h_j = Encoder(\mathcal{G}_j)
\end{equation}
, which is similar to Equation \ref{embedding}.
$h_i \in \mathcal{R}^{m \times F}, h_j \in \mathcal{R}^{n \times F}$ consists of $m$ and $n$ nodes, respectively. Referring Equation \ref{optimal_transport} and \ref{monge_map}, we can map each node in subgraph through Monge map as below:
\begin{equation}
\label{our_monge_map}
h'_i = B(p^*_i,h_i), \,\,h'_j = B(p^*_j, h_j)
\end{equation}
The $h'_i,h'_j \in \mathcal{R}^{C \times F}$ is the output of subgraph after mapping and aggregation. 
Though linear OT (Eq. \ref{linear_ot}) is also considerable, we choose the Monge Map which shows the better performance.

\textbf{Prediction}
Finally, using the concatenation of $h'_i,h'_j$ as an input of matching function M, we can estimate their similarity as below:
\begin{equation}
\label{edge_score}
w_{ij}=M(h'_i \oplus h'_j)
\end{equation}
If two inputs share the same label, the value of $w_{ij}$ should be closer to 1, and otherwise 0. We further employ node classification function $f(\cdot)$ (MLP) to predict the label of each subgraph's central node $h^e_i,h^e_j$, where $\mathcal{L}_{nll}$ is negative log-likelihood function:
\begin{equation}
\label{sm_loss}
\mathcal{L}_{SM}= \sum_{Y_i\neq Y_j}|w_{ij}| + \sum_{Y_i = Y_j}|w_{ij}-1| + \mathcal{L}_{nll}(f(h^e_i), Y_i) + \mathcal{L}_{nll}(f(h^e_j), Y_j).
\end{equation}

\begin{algorithm}
\caption{Confidence-based Subgraph Matching}
\label{pseudoPSO}
\begin{algorithmic}[1]
\Require Adjacency matrix $A$, initialized parameters $\theta_{SM}$, number of entire training epochs $K_e$, learning ratio $\eta$
\Ensure Supplementary edge coefficients $w$
\For{number of entire training epochs $K_e$}
\State Get embedding of nodes through Eq. \ref{embedding}
\State Get reference points $r$ by averaging embedding of training \par  nodes per each class
\State Find top-k confident edges through Eq. \ref{similarity}
\State Using the confident edges, reconstruct adjacency matrix \par within 2-hop as $\tilde{A}=A \lor A^2$
\State Sampling positive or negative node pair $(i,j)$
\For{each node pair $(i,j)$}
    \State Find the subgraphs $\mathcal{G}_i=\tilde{A}[i]$, $\mathcal{G}_j=\tilde{A}[j]$, and obtain \par \hspace\algorithmicindent their embedding $h_i,h_j$ through Eq. \ref{feat_trans}
    \State Apply Monge map for each subgraph Eq. \ref{our_monge_map} \par 
    \State Retrieve the edge coefficient through Eq. \ref{edge_score}
    \State Compute the loss $\mathcal{L}_{SM}$ using Eq. \ref{sm_loss} 
    \State Update $\theta'_{SM}=\theta_{SM}-\eta{\partial\mathcal{L}_{SM} \over \partial\theta_{SM}}$
\EndFor
\State{\textbf{end for}}
\EndFor
\State{\textbf{end for}}
\State Calculate supplementary edge coefficients $w$ using $\theta'_{SM}$
\end{algorithmic}
\end{algorithm}
Our subgraph matching module can be trained through Equation \ref{sm_loss}, and we describe the overall procedure in Algorithm \ref{pseudoPSO}.

\subsection{GNNs with Supplementary Edge Weights} \label{label_prop}
Recent studies focus on the strategy to better utilize edge weights. For example, some of them directly construct adjacency matrix \cite{velickovic2017graph, bo2021beyond,luo2021learning}, while others employ label propagation (LP) \cite{bui2018neural,stretcu2019graph,wang2020unifying} on GNNs to deal with uncertainty as below:
\begin{equation}
\label{vanilla_lp}
\mathcal{L}= \mathcal{L}_{GNN} + \lambda \mathcal{L}_{LP}.
\end{equation}
$\mathcal{L}_{GNN}$ is a widely used loss function for semi-supervised node classification (e.g., GCN \cite{kipf2016semi}) that is defined as follows:
\begin{equation}
\begin{gathered}
\widehat{Y}=softmax(\widehat{A} \sigma (\widehat{A}XW_0) W_1), \\
\mathcal{L}_{GNN}=\mathcal{L}_{nll}(Y, \widehat{Y}).
\end{gathered}
\end{equation}
, where $W$ is a learnable matrix. Though many recently proposed methods \cite{velickovic2017graph,xu2018powerful,klicpera2018predict,chen2020simple,bo2021beyond} are considerable for GNNs, here, we select GCN to show the efficacy of our method.
Back into Equation \ref{vanilla_lp}, $\lambda$ is a regularizer and $\mathcal{L}_{LP}$ gives additional penalties as below:
\begin{equation}
\label{original_lp}
\mathcal{L}_{LP}=\alpha_1 \sum_{i,j \in LL}w_{ij}d(i,j) + \alpha_2 \sum_{i,j \in LU}w_{ij}d(i,j) + \alpha_3 \sum_{i,j \in UU}w_{ij}d(i,j).
\end{equation}
The notation $\{L, U\}$ denotes labeled and unlabeled nodes, where $LU$ means that only a single node is labeled. $w_{ij} \in \{0, 1\}$ is a binary value that represents a connection between two nodes $i,j$, and $d$ is a dissimilarity measuring function (e.g., cosine similarity). Here, $\{a_1,a_2,a_3\}$ acts as a hyper-parameter. 
Recently, graph agreement model (GAM) \cite{stretcu2019graph} contemplates the limitation of fixed $w_{ij}$, and substitute it as a parameterized model $w_{ij}=g(X_i,X_j)$, where $g$ is a fully-connected networks. 
However, these methods implicate two limitations. Firstly, they have shown inferior performance for discriminating task-irrelevant edges under semi-supervised learning.
Secondly, the estimated $w_{ij}$ scales from zero to one, even making disassortative nodes similar (P-reg \cite{yang2021rethinking} also implicates this limitation). 
Thus, we improve Equation \ref{original_lp} as below:
\begin{equation}
\label{our_lp}
\begin{gathered}
\mathcal{L}_{SUP}=\alpha_1 \bigg( \sum_{i,j \in LU, w_{ij} > k}w_{ij}d(i,j) + \sum_{i,j \in LU, w_{ij} \leq k}(1-w_{ij})(1-d(i,j)) \bigg) \\ + \alpha_2 \bigg( \sum_{i,j \in UU, w_{ij} > k}w_{ij}d(i,j) + \sum_{i,j \in UU, w_{ij} \leq k}(1-w_{ij})(1-d(i,j)) \bigg).
\end{gathered}
\end{equation}
By sorting the score of entire edges $w$, we can retrieve a threshold $k$ based on a given confidence ratio. In Equation \ref{our_lp}, weights $w_{ij}$ that are greater than $k$ are trained to reduce dissimilarity $d(i,j)$, while others are guided to be dissimilar $1-d(i,j)$. Referring to Equation \ref{vanilla_lp}, we replace $\mathcal{L}_{LP}$ with our $\mathcal{L}_{SUP}$ and define $\mathcal{L}_G$ as below:
\begin{equation}
\label{gnn_loss}
\mathcal{L}_{G}= \mathcal{L}_{GNN} + \lambda \mathcal{L}_{SUP}.
\end{equation}
As described in GAM \cite{stretcu2019graph}, we exclude edges between labeled nodes ($i,j \in LL$) and set $\alpha_1=1.0$, $\alpha_2=0.5$. We set $0.01 \leq \lambda \leq 0.1$ which is proportional to the disassortativity of dataset.

\subsection{Optimization Strategy}
So far, we define losses of our subgraph matching with label propagation in Equation \ref{sm_loss} and \ref{gnn_loss}. Let us assume the parameters of the subgraph matching module $\theta_{SM}$ and the GNN module $\theta_{G}$ without sharing parameters.
Here, we notice that our ConSM implicates two limitations for optimization. Firstly, it is hard to determine whether the subgraph matching module $\theta_{SM}$ is converged or not. Secondly, the predicted edge coefficients may implicate uncertainty, which can impede the training of GNNs.
To solve this, in Algorithm \ref{alg_optim}, we suggest saving parameters $\theta'_{G}$ only if it attains the best validation score (line 13). Then, before we compute the loss of the next training sample, we can load these parameters if they exist (line 14).
Through this mechanism, we can guide GNNs to achieve better performance apart from the uncertainty of supplementary weights.

\begin{algorithm}
\caption{Overall Optimization of ConSM}
\label{alg_optim}
\begin{algorithmic}[1]
\Require Initialized parameters $\theta_{SM}$ and $\theta_{G}$, number of entire training epochs $K_e$, best validation score $\beta'=0$, learning ratio $\eta$
\Ensure Trained parameters $\theta'_{SM}$, $\theta'_{G}$
\For{number of entire training epochs $K_e$}
\For{training samples}
\State Compute the $\mathcal{L}_{SM}$ using Eq. \ref{sm_loss}
\State Update $\theta'_{SM}=\theta_{SM}-\eta{\partial \mathcal{L}_{SM} \over \partial \theta_{SM}}$
\EndFor
\State{\textbf{end for}}
\State Retrieve edge coefficients $w$ using $\theta'_{SM}$
\For{training samples}
\State Compute the $\mathcal{L}_{G}$ using Eq. \ref{gnn_loss}
\State Compute the validation score $\beta$
\State Update $\theta'_{G}=\theta_{G}-\eta{\partial \mathcal{L}_{G} \over \partial \theta_{G}}$
\If {$\beta$ > $\beta'$}
\State Save updated parameters  $\theta'_{G}$
\State $\beta'=\beta$
\EndIf
\EndFor
\State{\textbf{end for}}
\State Load $\theta'_{G}$ if exists
\EndFor
\State{\textbf{end for}}
\end{algorithmic}
\end{algorithm}

\begin{table}[ht]
\caption{Statistical details of homophilic datasets}
\label{homo_dataset}
\vspace{-3mm}
\centering
\begin{center}
\begin{tabular}{lllll}
\multicolumn{1}{l}{}    & \multicolumn{1}{l}{}    &        &         &                \\ 
\Xhline{2\arrayrulewidth}
        & Datasets                       & Cora & Citeseer & Pubmed \\ 
\Xhline{2\arrayrulewidth}
                        & \# Nodes                       & 2,708  & 3,327   & 19,717\\
                        & \# Edges               & 10,558  & 9,104  & 88,648 \\
                        & \# Features             & 1,433  & 3,703  & 167,597 \\
                        & \# Classes             & 7  & 6  & 3       \\
                        & \# Training Nodes            & 140  & 120  & 60  \\
                        & \# Validation Nodes            & 1,568  & 2,207  & 18,657  \\
                        & \# Test Nodes                 & 1,000  & 1,000  & 1,000       \\
\Xhline{2\arrayrulewidth}
\end{tabular}
\end{center}
\end{table}

\begin{table}[ht]
\caption{Statistical details of heterophilic datasets}
\label{hetero_dataset}
\centering
\begin{center}
\begin{tabular}{lllll}
\multicolumn{1}{l}{}    & \multicolumn{1}{l}{}    &        &         &  \\ 
\Xhline{2\arrayrulewidth}
        & Datasets                       & Actor & Chameleon & Squirrel  \\ 
\Xhline{2\arrayrulewidth}
                        & \# Nodes                       & 7,600 & 2,277  & 5,201   \\
                        & \# Edges               & 25,944 & 33,824  & 211,872  \\
                        & \# Features            & 931 & 2,325  & 2,089  \\
                        & \# Classes             & 5  & 5  & 5       \\
                        & \# Training Nodes            & 100  & 100  & 100  \\
                        & \# Validation Nodes              & 3,750 & 1,088  & 2,550  \\
                        & \# Test Nodes                 & 3,750 & 1,089  & 2,551      \\
\Xhline{2\arrayrulewidth}
\end{tabular}
\end{center}
\end{table}

\subsection{Computational Complexity Analysis}
The computational costs of our model can be divided into two parts. The first one is a vanilla GCN \cite{kipf2016semi} model whose complexity is known as $\mathcal{O}(|\mathcal{E}|P_{GCN})$, where they are proportional to the number of entire edges $| \mathcal{E} |$ and the size of learnable matrices $P_{GCN}$. The second term is our ConSM which computes the similarity between two subgraphs. Instead of naively calculating Wasserstein distance $\mathcal{O}(n^3log(n))$, we conduct linear mapping \cite{kolouri2020wasserstein} and measure Euclidean distance, which can be retrieved through simple matrix multiplication. Consequently, our computational cost can be defined as $\mathcal{O}(|\mathcal{E}|P_{GCN} + |\mathcal{E}_M|P_{ConSM})$, where $|\mathcal{E}_M|$ stands for the number of edges included in sampled subgraph, and $P_{ConSM}$ is the set of parameters in subgraph matching module.

\section{EXPERIMENTS} \label{experiments}
In this section, we compare our ConSM with several state-of-the-art methods using a homophilic and heterophilic graph dataset. In particular, we aim to answer the following research questions:
\begin{itemize}
    \item \textbf{RQ1:} Does ConSM improves node classification accuracy compared to the state-of-the-art approaches? 
    \item \textbf{RQ2:} How much does ConSM accurately specify task-irrelevant edges in terms of graph denoising?
    \item \textbf{RQ3:} Does the confidence ratio for the subgraph matching module affects the overall classification result?
    \item \textbf{RQ4:} Can ConSM alleviates over-smoothing for stacking many layers effectively?
\end{itemize}

\subsection{Dataset Description and Baselines}
\textbf{Dataset description} We conduct investigations with the following publicly available dataset. The statistical details are described in Table \ref{homo_dataset} and \ref{hetero_dataset}, where we categorize them into two types; assortative and disassortative networks. The explanations of each dataset are demonstrated below.

\begin{itemize}
    \item \textbf{Assortative networks} For assortative data, we adopt widely used benchmark graphs; Cora, Citeseer, and Pubmed \cite{kipf2016semi}. Here, each node represents a paper and the edge denotes a citation between two papers. Node features stand for the bag-of-words of paper, and each node has a unique label based on its relevant topic.
    \item \textbf{Disassortative networks}
    We adopt Actor co-occurrence graph \cite{tang2009social} and Wikipedia network \cite{rozemberczki2019gemsec} as disassortative graphs. For Actor co-occurrence data, the node stands for an actor, and the edges are co-occurrence on the same Wikipedia pages. The node label denotes five types based on the keywords of an actor. Similarly, the Wikipedia network consists of Chameleon and Squirrel, where the edges are hyperlinks between web pages. The node features are 
    several informative nouns and we classify them into five categories based on their monthly traffic.
\end{itemize}

\textbf{Baselines}
Using the above datasets, we compare our method with the state-of-the-art baselines.
A brief explanation of these methods can be seen as follows:

\begin{table}
\caption{Node classification accuracy (\%) on homophilic citation networks. Bold$^\ast$ symbol indicates the best performance, and methods with $\dag$ are built upon GCN.}
\label{perf_1}
\centering
\begin{center}
\begin{adjustbox}{width=0.48\textwidth}
\begin{tabular}{lllll}
&  & & &              \\ 
\Xhline{2\arrayrulewidth}
        & Datasets                       & Cora & Citeseer & Pubmed \\ 
        & Hom. ratio ($h$) & 0.81 & 0.74 & 0.8 \\
\Xhline{2\arrayrulewidth}
                        & MLP              & 54.2 $_{\,\pm\,0.5\,\%}$ & 53.7 $_{\,\pm\,1.7\,\%}$ & 69.7 $_{\,\pm\,0.4\,\%}$ \\
                        & GCN               & 80.5 $_{\,\pm\,0.4\,\%}$  & 67.5 $_{\,\pm\,0.6\,\%}$ & 78.4 $_{\,\pm\,0.2\,\%}$ \\
                        & DropEdge$^\dag$             & 80.6 $_{\,\pm\,0.4\,\%}$ & 68.4 $_{\,\pm\,0.5\,\%}$ & 78.3 $_{\,\pm\,0.3\,\%}$ \\
                        & GAT             & 81.4 $_{\,\pm\,0.4\,\%}$ & 69.3 $_{\,\pm\,0.9\,\%}$ & 78.6 $_{\,\pm\,0.5\,\%}$ \\
                        & GIN              & 78.2 $_{\,\pm\,0.2\,\%}$ & 65.1 $_{\,\pm\,0.6\,\%}$ & 76.8 $_{\,\pm\,0.8\,\%}$ \\
                        & APPNP                 & 82.1 $_{\,\pm\,0.4\,\%}$ & 69.2 $_{\,\pm\,0.7\,\%}$ & 78.7 $_{\,\pm\,0.4\,\%}$ \\
                        & GCNII           & 81.1 $_{\,\pm\,0.3\,\%}$ & 67.0 $_{\,\pm\,0.4\,\%}$ & 78.5 $_{\,\pm\,0.8\,\%}$ \\
                        & GAM$^\dag$          & 81.3 $_{\,\pm\,0.6\,\%}$ & 70.4 $_{\,\pm\,0.3\,\%}$ & 79.2 $_{\,\pm\,0.1\,\%}$ \\
                        & H$_2$GCN        & 80.5 $_{\,\pm\,0.2\,\%}$ & 68.9 $_{\,\pm\,0.5\,\%}$ & 78.9 $_{\,\pm\,0.2\,\%}$ \\
                        & FAGCN        & 81.5 $_{\,\pm\,0.5\,\%}$ & 68.6 $_{\,\pm\,0.2\,\%}$ & 79.0 $_{\,\pm\,0.1\,\%}$ \\
                        & PTDNet$^\dag$      & 81.5 $_{\,\pm\,0.7\,\%}$ & 69.4 $_{\,\pm\,0.6\,\%}$ & 76.9 $_{\,\pm\,1.1\,\%}$ \\
\Xhline{2\arrayrulewidth}
                        & Ours$^\dag$      & \textbf{83.6$^*$ $_{\,\pm\,0.3\,\%}$} & \textbf{71.2$^*$ $_{\,\pm\,0.2\,\%}$} & \textbf{79.6$^*$ $_{\,\pm\,0.1\,\%}$} \\
\Xhline{2\arrayrulewidth}
\end{tabular}
\end{adjustbox}
\end{center}
\end{table}

\begin{itemize}
    \item \textbf{MLP} \cite{popescu2009multilayer} employs a feed-forward neural network that only utilizes a central node for classification.
    \item \textbf{GCN} \cite{kipf2016semi} is a traditional GNN models that suggests first order approximation of Chebyshev polynomials \cite{defferrard2016convolutional} to localize spectral filters.
    \item \textbf{DropEdge} \cite{rong2019dropedge} randomly removes edges under a given probability to alleviate over-fitting problem.
    \item \textbf{GAT} \cite{velickovic2017graph} specifies different weights between two nodes, while ignoring graph Laplacian matrix.
    \item \textbf{GIN} \cite{xu2018powerful} pointed out the limited discriminative power of GCN, suggesting a graph isomorphism network that satisfies the injectiveness condition.
    \item \textbf{APPNP} \cite{klicpera2018predict} combines personalized PageRank with GCN that improves prediction accuracy, while reducing computational complexity.
    \item \textbf{GCNII} \cite{chen2020simple} integrates identity mapping to redeem the deficiency of APPNP. 
    \item \textbf{GAM} \cite{stretcu2019graph} adopts the graph agreement model under the assumption that not all edges correspond to sharing the same label between nodes. 
    \item \textbf{H$_2$GCN} \cite{zhu2020beyond} suggests ego-neighbor separation and hop-based aggregation to deal with heterophilic graph.
    \item \textbf{FAGCN} \cite{bo2021beyond} further utilizes high-frequency signal beyond low-frequency information in GNNs.
    \item \textbf{PTDNet} \cite{luo2021learning} proposes a topological denoising network to prune task-irrelevant edges as a downstream task of GNNs.
\end{itemize}

\begin{table}
\caption{Node classification accuracy (\%) on heterophilic Wikipedia pages. Bold$^\ast$ symbol indicates the best performance, and methods with $\dag$ are built upon GCN.}
\label{perf_2}
\centering
\begin{center}
\begin{adjustbox}{width=0.48\textwidth}
\begin{tabular}{lllll}
  &     &        &         &               \\ 
\Xhline{2\arrayrulewidth}
        & Datasets                       & Actor & Chameleon & Squirrel \\ 
        & Hom. ratio ($h$) & 0.22 & 0.23 & 0.22 \\
\Xhline{2\arrayrulewidth}
                        & MLP                       & \textbf{27.9$^*$ $_{\,\pm\,1.1\,\%}$} & 41.2 $_{\,\pm\,1.8\,\%}$ & 26.5 $_{\,\pm\,0.6\,\%}$ \\
                        & GCN               & 21.4 $_{\,\pm\,0.6\,\%}$  & 49.4 $_{\,\pm\,0.7\,\%}$ & 33.1 $_{\,\pm\,1.2\,\%}$ \\
                        & DropEdge$^\dag$           & 21.9 $_{\,\pm\,0.4\,\%}$ & 49.2 $_{\,\pm\,0.8\,\%}$ & 31.0 $_{\,\pm\,1.4\,\%}$ \\
                        & GAT             & 23.2 $_{\,\pm\,0.8\,\%}$ & 48.2 $_{\,\pm\,0.6\,\%}$ & 30.1 $_{\,\pm\,0.9\,\%}$ \\
                        & GIN             & 23.6 $_{\,\pm\,0.5\,\%}$ & 40.1 $_{\,\pm\,3.7\,\%}$ & 23.0 $_{\,\pm\,2.4\,\%}$ \\ 
                        & APPNP                 & 21.7 $_{\,\pm\,0.2\,\%}$ & 45.2 $_{\,\pm\,0.7\,\%}$ & 30.6 $_{\,\pm\,0.7\,\%}$ \\
                        & GCNII           & 25.7 $_{\,\pm\,0.4\,\%}$ & 45.1 $_{\,\pm\,0.5\,\%}$ & 28.8 $_{\,\pm\,0.3\,\%}$ \\
                        & GAM$^\dag$             & 21.7 $_{\,\pm\,0.3\,\%}$ & 49.5 $_{\,\pm\,0.5\,\%}$ & 33.9 $_{\,\pm\,0.4\,\%}$ \\
                        & H$_2$GCN        & 22.8 $_{\,\pm\,0.3\,\%}$  & 46.6 $_{\,\pm\,1.2\,\%}$ & 29.8 $_{\,\pm\,0.7\,\%}$ \\
                        & FAGCN        & 26.9 $_{\,\pm\,0.4\,\%}$  & 46.7 $_{\,\pm\,0.6\,\%}$  & 29.5 $_{\,\pm\,0.7\,\%}$ \\
                        & PTDNet$^\dag$        & 21.5 $_{\,\pm\,0.5\,\%}$ & 49.7 $_{\,\pm\,0.9\,\%}$ & 32.6 $_{\,\pm\,0.7\,\%}$ \\
\Xhline{2\arrayrulewidth}
                        & Ours$^\dag$        & 27.7 $_{\,\pm\,0.4\,\%}$  & \textbf{52.3$^*$ $_{\,\pm\,0.4\,\%}$} & \textbf{35.7$^*$ $_{\,\pm\,0.5\,\%}$} \\
\Xhline{2\arrayrulewidth}
\end{tabular}
\end{adjustbox}
\end{center}
\end{table}

\subsection{Experimental Setup} 
All methods are implemented in \textit{PyTorch Geometric}\footnote{\label{footnote}https://pytorch-geometric.readthedocs.io/en/latest/modules/nn.html}, with Adam optimizer (weight decay $5e^{-4}$) and proper learning ratio ($1e^{-3}$). We set the embedding dimension as 64 for all methods, but diversifying it can improve the overall performance \cite{luo2021graph}. Here, we adopt 2 layers of GNNs for all baselines, while APPNP, GCNII, and GIN further utilize 2 layers of fully-connected networks for classification.
We apply ReLU as an activation function except for PTDNet (Sigmoid is used here). The Softmax is applied on the last hidden layers for classification. For all datasets, we randomly select 20 samples per class as a training set, and the rest is for validation and testing (please refer to Table \ref{homo_dataset} and \ref{hetero_dataset}). The performance is evaluated based on a test set accuracy that achieved the best validation score.

\subsection{Results and Discussion (RQ1)}
In Table \ref{perf_1} and \ref{perf_2}, we describe experimental results of baselines and our method that are conducted under homophilic / heterophilic datasets. Here, let us assume the homophily ratio $h$ as below:
\begin{equation}
h = {\text{\# of edges that connect nodes with same label} \over \text{\# of entire edges}}
\end{equation}

\textbf{Results on homophilic graph datasets} In Table \ref{perf_1}, we first discuss performance on three homophilic datasets, where most of the connected nodes share the same label. 
We conduct experiments over 10 times and report the mean and variance of test accuracy. We also describe the performance of MLP to show the influence of message passing on graph datasets.
Firstly, for methods that employ GCN as a backbone (marked with $\dag$), our approach achieves state-of-the-art performance on multiple benchmark datasets. Specifically, our method outperforms GCN over 3.7 \%, 5.2 \%, 1.6 \%, respectively. Among baselines, in \textit{Citeseer}, DropEdge shows better performance than GCN which has relatively low homophily than other networks. Above all, APPNP and GAM achieve the best performance with the aid of label propagation, followed by GAT adopting an attention mechanism. For our experiments, GCNII shows lower performance than APPNP, which means that emphasizing the identity feature is not suitable for homophilic data. The design choice of rest algorithms (H$_2$GCN, FAGCN, PTDNet) are for heterophilic graphs, where they fail to achieve notable improvements over GCN.

\begin{figure}
  \includegraphics[width=0.48\textwidth]{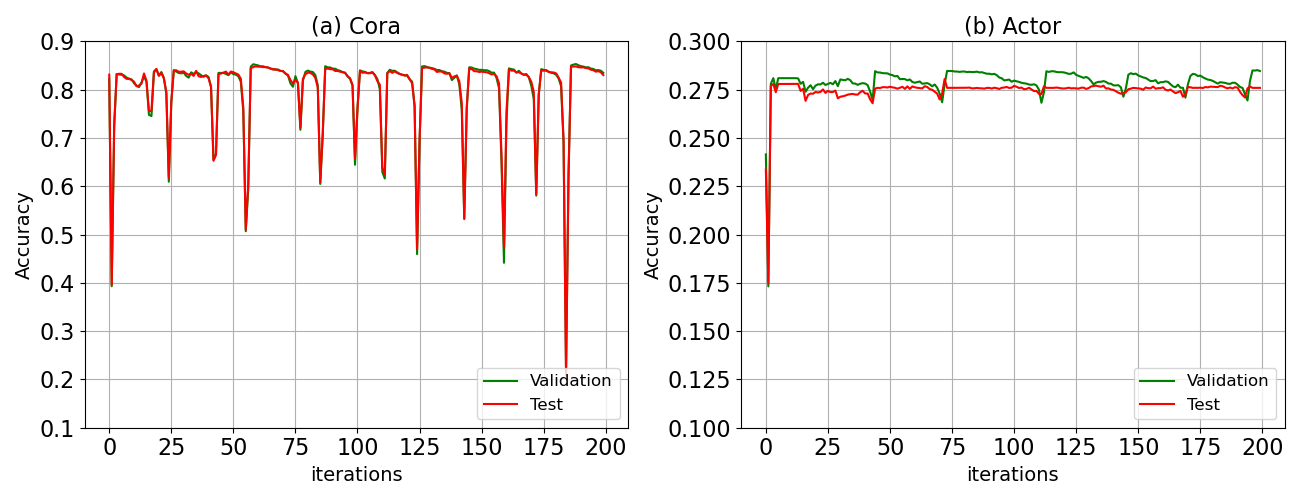}
  \caption{Convergence analysis on (a) Cora, and (b) Actor. Each figure contains validation (green) and test (red) accuracy of node classification}
  \label{convergence}
\end{figure}

\textbf{Results on heterophilic graph datasets}
In addition to the homophilic network, we conduct the experiments under heterophilic data with the same settings and plot the results in Table \ref{perf_2}. As can be seen, these graphs are generally disassortative with a low homophilic ratio $h$, which can impede the advantages of message passing in GNNs. Surprisingly, MLP achieves the best performance for \textit{Actor}, followed by our ConSM, FAGCN, and GCNII. Given that GCNII outperforms APPNP, we guess that a central node is highly important for the $Actor$ network. Nonetheless, these methods fail to outperform GCN for different datasets.
Instead, our method achieves the best accuracy on both $Chameleon$, and $Squirrel$. Based on the results that GAM shows outstanding performance for this kind of network, the supplementary model generalizes well under a heterophilic structured dataset. Under our experiments, H$_2$GCN, FAGCN, and PTDNet have shown to achieve lower scores, which will be discussed in Section \ref{sec_rq2}.

\textbf{Convergence analysis}
To better understand the convergence of ConSM, in Figure \ref{convergence}, we describe validation and test accuracy for training. The x-axis illustrates iterations, while the y-axis is classification accuracy. As described in Algorithm \ref{alg_optim}, a single iteration is a combination of training subgraph matching modules, followed by training GNN layers.
Here, the validation and test accuracy vary significantly, but ConSM manages to achieve better performance as iteration increases. This is because ConSM loads parameters of the best validation score (please refer to Algorithm \ref{alg_optim}), which can prevent the uncertainty of supplementary information precisely.

\subsection{Edge Classification (RQ2)} \label{sec_rq2}
To validate whether ConSM can predict edge coefficients correctly, we examine the accuracy of our method and several state-of-the-art approaches. Here, we assume the label of edges that connect two nodes with the same class as 1 and vice versa. 

\begin{figure}
  \includegraphics[width=0.48\textwidth]{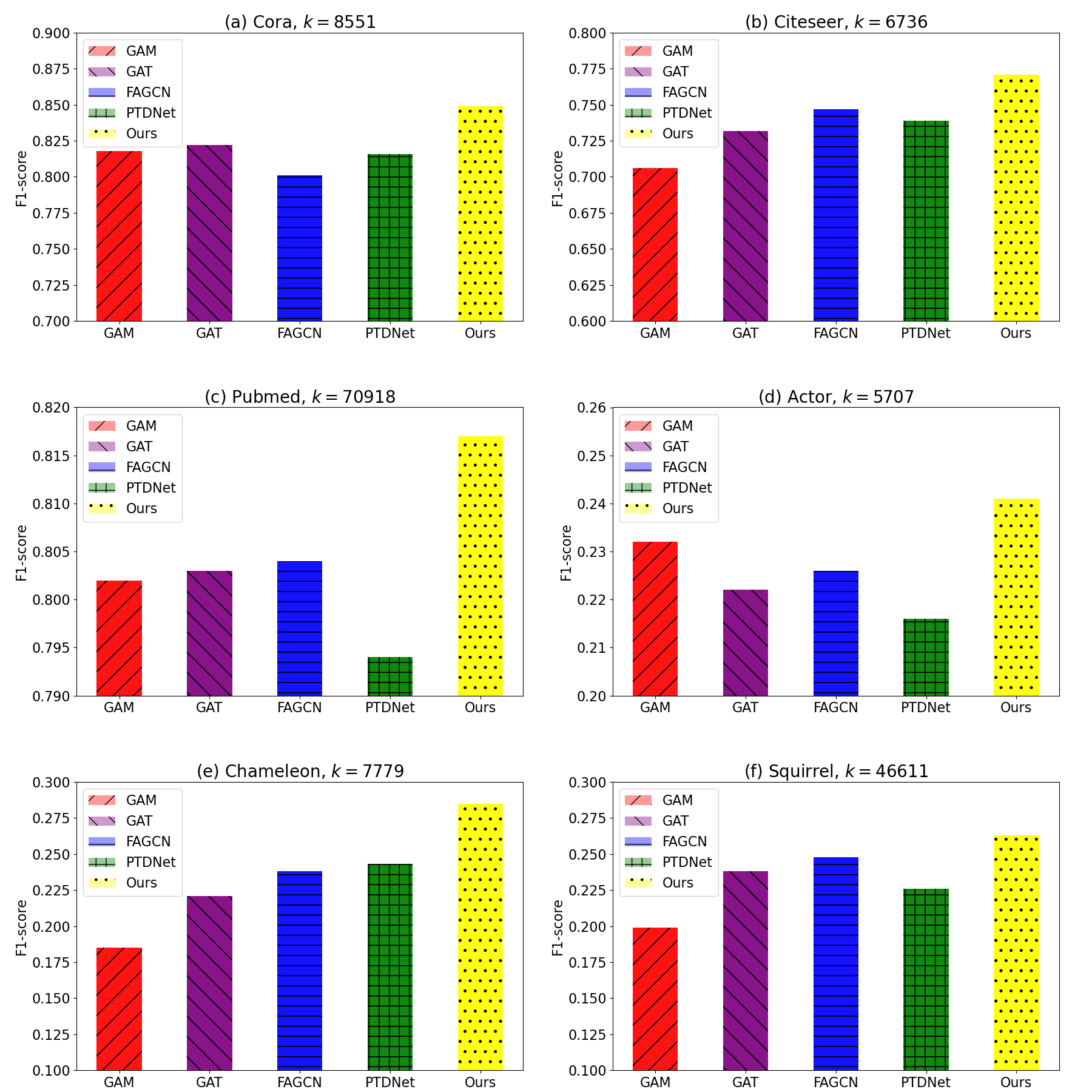}
  \caption{We measure F1-score to evaluate edge classification performance on six graph datasets. Here, we adopt our model with four baselines that specify edge coefficients}
  \label{edge_clf}
  \vspace{-2mm}
\end{figure}

For each method, we sort their predicted coefficients and select $k-th$ largest value as a threshold, which is equal to the number of positive edges.  
Specifically, for (a) Cora, $h=0.81$ and $\mathcal{E}=10,558$ (please refer Table \ref{homo_dataset} and \ref{perf_1}), and thus, $k=\lfloor10,558 \times 0.81\rfloor$ which is described in Figure \ref{edge_clf}. We adopt F1-score that has shown to be effective for binary classification as below:
\begin{equation}
    \text{F1-score}={2 \times \text{precision} \times \text{recall} \over \text{precision} + \text{recall}}
\end{equation}

We first introduce some details of baselines, followed by a discussion on the experimental results.
(1) GAM: as described in Equation \ref{gam}, the agreement model generates the same class probability. We employ their edge coefficients with the best validation result.
(2) GAT: we exclude node-wise normalization (e.g., softmax), which can be highly sensitive to the degree of central nodes. Then, using the representations of the final hidden layer, we retrieve the attention value of the entire edges. Multi-head attention is applied for the front layers, while the final layer only employs single-head attention. 
(3) FAGCN: they retrieve the coefficients following the Equation \ref{fagcn}. Similar to GAT, we exploit the attention values of the last hidden representations. The hyper-parameters are tuned referring \cite{bo2021beyond}.
(4) PTDNet: similar to previous studies, we adopt the generated graphs using final representations of GNNs. The hyper-parameters are remain the same as their \textit{implementations}\footnote{https://github.com/flyingdoog/PTDNet}.
(5) Ours: the coefficients of our model can be retrieved through Equation \ref{edge_score}.

In Figure \ref{edge_clf}, we can see that GAM shows the lowest performance for most graph datasets, except for (d) Actor. It is not surprising since they only utilize a central node for a prediction. Here, GAT relatively outperforms GAM with the aid of message passing and attention layer. Except for (a) Cora, FAGCN achieves better performance than GAT, which describes the effectiveness of high-frequency signals. 
Notably, PTDNet is not shown to be powerful enough, where the edge pruning between communities fails to generalize on most graph datasets. Comparatively, our model improves the F1-score significantly for all datasets, which justifies the necessity of confidence-aware subgraph matching.

\begin{figure}
  \includegraphics[width=0.48\textwidth]{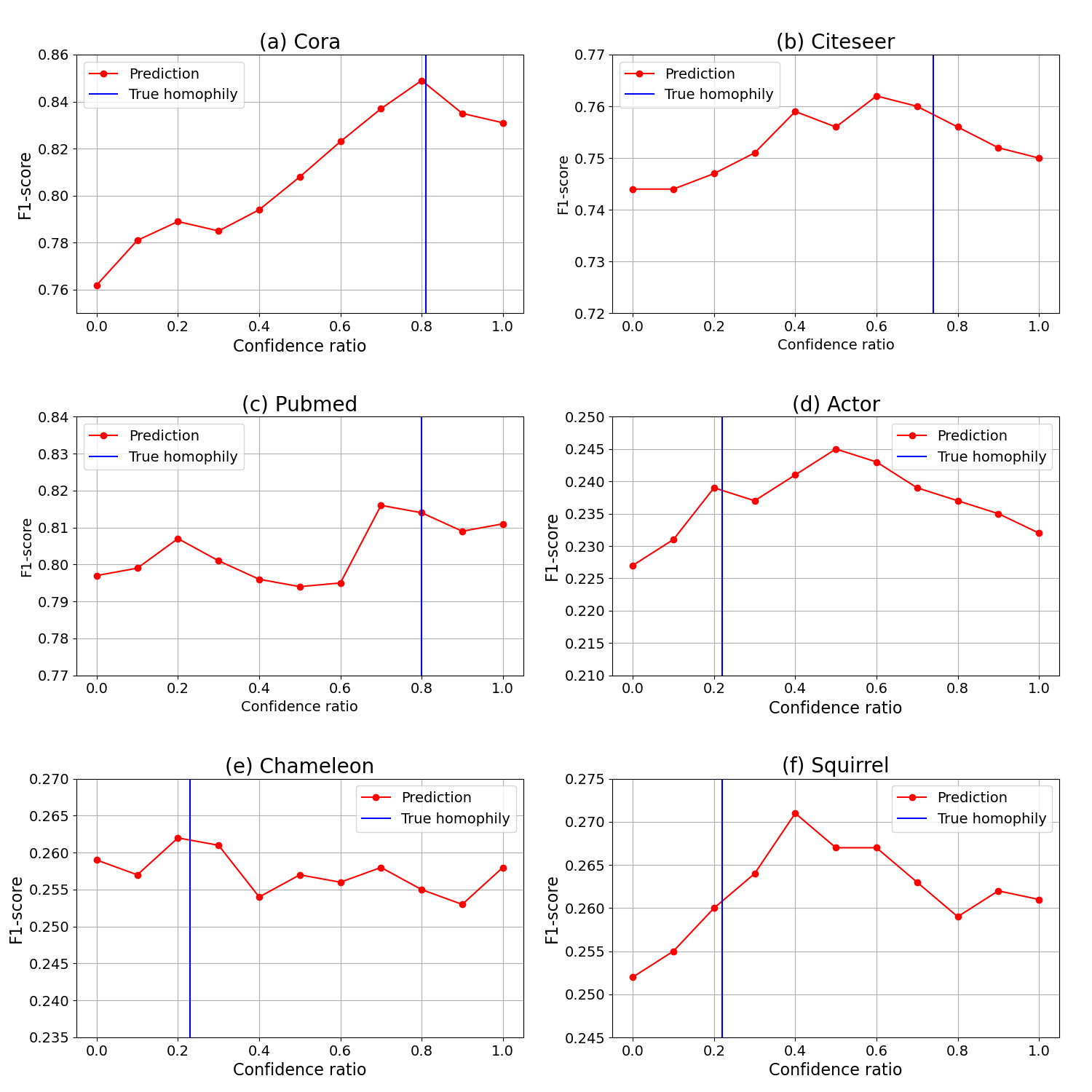}
  \caption{We differentiate the confidence ratio of subgraph matching module, and describe F1-score on six graph datasets}
  \label{param_sense}
\end{figure}

\subsection{Parameter Sensitivity Analysis (RQ3)}
In this section, we further measure edge classification scores by differentiating a hyper-parameter of the subgraph matching module. To deal with heterophily, we introduced a confidence ratio ($\zeta$) to reflect data homophily, assuming that connected nodes may not share the same labels. In Figure \ref{param_sense}, we plot F1-score on six datasets by varying $\zeta$ from 0 to 1. We also describe true homophily ratio (please refer $h$ in Table \ref{perf_1} and \ref{perf_2}) as blue lines. If $\zeta=0$, the supplementary module does not utilize neighboring nodes for a prediction, while $\zeta=1$ means that it fully utilizes adjacent nodes.

Here, a confidence ratio ($\zeta$) that shows the best F1-score fairly aligns well with the true homophily ratio $h$, and the selection of $\zeta$
is important for precise prediction. Though $\zeta=0.5$ is quite different from $h=0.22$ for (d) Actor, we insist that disassortative neighbors can also contribute to improving classifications, as we described in Figure \ref{example_2}.
Nonetheless, we admit that a choice of $\zeta$ is quite sensitive, and may require human efforts to achieve the best accuracy.

\subsection{Analysis on Over-smoothing (RQ4)}
\begin{figure}
  \includegraphics[width=0.48\textwidth]{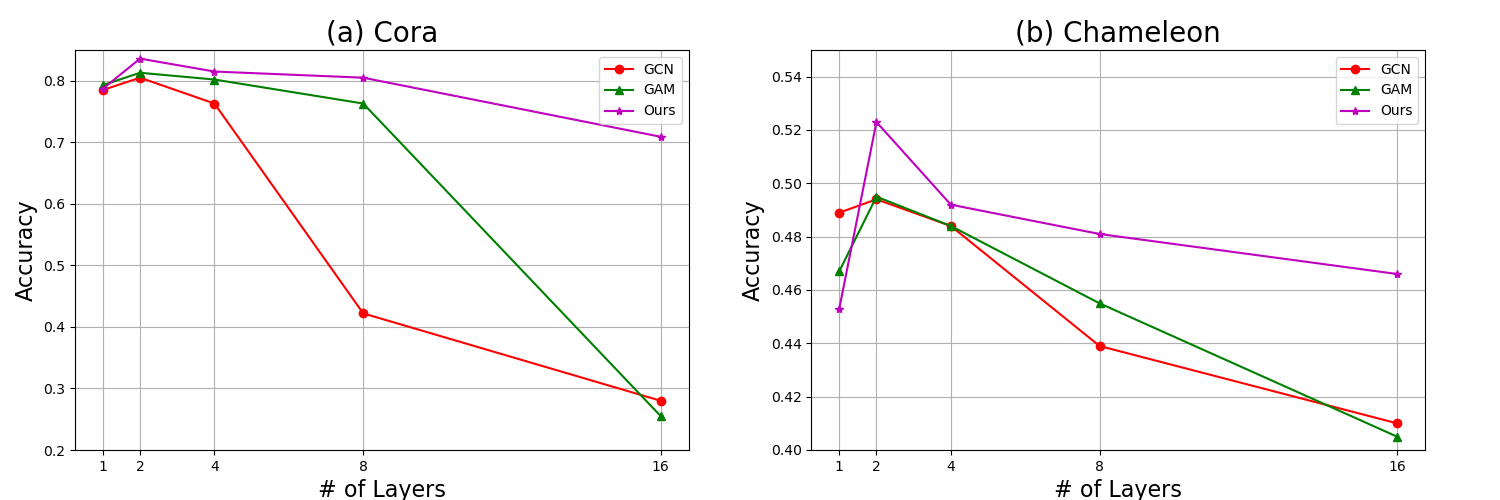}
  \caption{Evaluation on over-smoothing using (a) Cora, and (b) Chameleon dataset. We plot the accuracy of two baselines and our method using different number of layers}
  \label{smooth}
  \vspace{-3mm}
\end{figure}

Over-smoothing is a fundamental problem for GNNs when stacking multiple layers \cite{li2018deeper,zhao2019pairnorm}. Here, we scrutinize this phenomenon by differentiating the depth of layers as \{1, 2, 4, 8, 16\}, and report the node classification accuracy on (a) Cora, and (b) Chameleon. In Figure \ref{smooth}, we describe the results of GCN, GAM, and our ConSM. GCN shows the best performance at 2 layers on both datasets. However, they degrade slightly at 4 layers and dramatically decrease beyond it. This means that GCN itself cannot alleviate the over-smoothing problem. Though GAM remains relatively stable compared to GCN, they also suffer from smoothing when stacking more layers. Comparatively, ConSM consistently achieves the best performance, and the accuracy does not decrease severely for deeper layers (e.g., 16 layers).
We suggest that the integration of well-classified edge coefficients with label propagation effectively controls this problem, which shows the effectiveness of our method.

\section{CONCLUSION}
In this work, we suggest a confidence ratio to deal with multiple disassortative edges for semi-supervised node classification. We pointed out the significance of configuring edge weights precisely, and thus, we propose to measure similarity between two connected nodes using their subgraphs.
Further, based on the observations that directly applying the predicted weights are highly risky, we integrate label propagation with our confidence ratio to secure robustness and improve the overall performance. The extensive experiments for both homophilic and heterophilic setups well describe the superiority of our model.

\subsubsection*{\normalfont{\textbf{Acknowledgments}}}
This work was supported by the National Research Foundation of Korea (NRF) (No. 2016R1A5A1012966, No. 2020R1A2C110168713),  Institute of Information \& communications Technology Planning \& Evaluation (IITP) (No. 2021-0-02068 Artificial Intelligence Innovation Hub, No. RS-2022-00156287 Innovative Human Resource Development for Local Intellectualization support program) grant funded by the Korea government (MSIT).

\bibliographystyle{ACM-Reference-Format}
\bibliography{references.bib}

\end{document}